# A Bi-population Particle Swarm Optimizer for Learning Automata based Slow Intelligent System

Mohammad Hasanzadeh Mofrad, and S. K. Chang
School of Computing & Information
Department of Computer Science
University of Pittsburgh
Pittsburgh, USA
{MOHAMMAD.HMOFRAD and SCHANG}@pitt.edu

*Abstract*—Particle Swarm Optimization (PSO) is an Evolutionary Algorithm (EA) that utilizes a swarm of particles to solve an optimization problem. Slow Intelligence System (SIS) is a learning framework which slowly learns the solution to a problem performing a series of operations. Moreover, Learning Automata (LA) are minuscule but effective decision making entities which are best suited to act as a controller component. In this paper, we combine two isolate populations of PSO to forge the Adaptive Intelligence Optimizer (AIO) which harnesses the advantages of a bi-population PSO to escape from the local minimum and avoid premature convergence. Furthermore, using the rich framework of SIS and the nifty control theory that LA derived from, we find the perfect matching between SIS and LA where acting slowly is the pillar of both of them. Both SIS and LA need time to converge to the optimal decision where this enables AIO to outperform standard PSO having an incomparable performance on evolutionary optimization benchmark functions.

*Keywords—Particle Swarm Optimization; Slow Intelligence System; Learning Automata*

I. INTRODUCTION

Swarm Intelligence (SI) [1] brings a new breed of algorithms to the EA's community. Inspiring from the collective behavior of group of animals scavenging for food sources, the SI algorithms find their application in Artificial Intelligence (AI), Machine Learning (ML), social networks, Grid and Cloud computing and computer networks. SI is widely used for black box/white box optimization problems. Also, it is impressively useful for adjusting/adapting sensitive parameters of arbitrary data models.

The PSO [2] is a SI algorithm which is derived from the group movement of animal herds especially birds. PSO algorithm repeatedly try to guess the next feasible solution using its current information about each individual's best position and the population's best position heretofore. In PSO, the position of particles are updated using the velocity formula which contains the current position, personal best position of a particle and global best position of the swarm. Calculating the distance of each particle from its personal best position and global best position of the swarm, in each iteration of PSO algorithm, the particles move toward the latest optimal position of the swarm. This individual and social moves of each particle will eventually lead to finding the optimal result of the candidate problem.

A SIS [3] is a general framework that consists of a set of slow or quick decision cycles. In each decision cycle, SIS tries to search for new solutions within the problem space by applying a set of operators including *enumeration*, *propagation*, *adaptation*, *elimination* and *concentration*. In general, SIS is a slow learner but it gains performance over time by applying a sequence of operators to the problem space and producing the solution space.

LA [4] are autonomous machines designed for learning the optimal action within an unknown environment. LA roots in control theory where centralized or decentralized or even a mixture of both of these modes are used to study the behavior of a dynamic system with inputs and show how positive or negative feedback can modify the system's behavior. LA has application in AI [5]–[7], ML [8], EA[9], [10], distributed systems [11], [12], and image processing [13].

The AIO is a new SI-based optimization algorithm which used two isolated populations of PSO. In AIO, we follow the TDR concept [14] of SIS to break the problem dimensions into three sub-dimensions that each of them is controlled by a learning automaton. The PSO populations share information through producing the reinforcement signal for the LA that control which population to run on the specific sub-dimension of the problem space. It also utilizes the slow and quick decision cycles of SIS to adopt the inertia weight parameter of PSO.

The rest of this paper is laid out organized as follows: In Section II, we discuss the related work including PSO, SIS, and LA. In Section III, we present the proposed optimization algorithm AIO. We present the experimental results in Section IV and we conclude the paper in Section V.

II. RELATED WORK

*A. Particle Swarm Optimization (PSO)*

PSO [15] is an optimization algorithm inspires from the movement of flock of birds which often moves under the guidance of an individual leader bird to find nearby food. In PSO, a bird in the flock is simulated as a particle in the population. The population's best position and particle's best position mimic the leader bird and the best aviation position of each individual bird in terms of the provisioned food resource.

In an *n* dimensional space, the $i_{th}$ PSO's individual is attributed as follows:

- A *position* vector $X_i = (x_i^1, x_i^2, ..., x_i^n)$
- A *velocity* vector $V_i = (v_i^1, v_i^2, ..., v_i^n)$
- A *pbest* vector $pbest_i = (pbest_i^1, pbest_i^2, ..., pbest_i^n)$

In each iteration, the $X_i$ and $V_i$ are updated using the following equations:

$$V_{i+1} = w V_i + c_1 r_1 (pbest_i - X_i) + c_2 r_2 (gbest - X_i) \quad (1)$$

$$X_{i+1} = X_i + V_i \quad (2)$$

where in ( 1 ) and ( 2 ):

- *gbest* is the best global position of the population is $gbest_i = (gbest_i^1, gbest_i^2, ..., gbest_i^n)$
- $c_1$ & $c_2$ are acceleration constants
- $r_1$ & $r_2 \in [0, 1]$ are two random numbers
- *w* is the inertia weight

*B. Slow Intelligence System (SIS)*

The SIS [16] is a slow learner with multiple decision cycles. In each decision cycle a set of operations are applied to the existing solutions of the target problem. In a SIS Abstract Machine, these operations could be any combination of *Enumeration*, *Propagation*, *Adaptation*, *Elimination*, and *Concentration* operators. In each decision cycle of SIS, a predicate that is constructed from these operators is shielded by a *guard* operator which controls the flow of operation from computationally inexpensive decision cycles or *quick decision cycles* to expensive decision cycles or *slow decision cycles.*

SIS consists of the following steps:

- **Enumeration** of the different available solutions until finding the optimal solution
- **Propagation** of the achieved new information from the new solutions within a body of feasible solutions.
- **Adaptation** of the current solutions using the effective information gained from the elite solutions.
- **Elimination** of the worst solutions that exist in the problem space.
- **Concentration** on the elite solutions to produce new promising solutions.

An Abstract Machine for SIS in the $n_{th}$ decision cycle is defined as $M = [P, S, C]$ where:

- $P = \{p_0, p_1, p_2, ..., p_n\}$ is the problem space where $p_0$ is the initial problem set
- $S = \{s_1, s_2, ..., s_n\}$ is the solution space
- $C = \{cycle_1, cycle_2, ..., cycle_n\}$ is the computation cycle

Each SIS's decision cycle $cycle_n$ consists of a sequence of SIS's operators that are applied to the problem $p_n$ and creates a new solution $s_n$.

*C. Learning Automata (LA)*

LA [17] are probabilistic decision making elements. Having a series of interactions with the environment, they adopt to the environment iteratively and learn the optimal action. A widespread type of LA is Variable Structure Learning Automaton (VLSA). The $n_{th}$ step of a VLSA is defined by a quadruple $[P, α, β, T]$ where:

- $α = \{α_1, α_2, ..., α_r\}$ is the set of actions where *r* is the number of actions
- $β = \{β_1, β_2, ..., β_r\}$ is the set of inputs where in a stationary environment $β \in \{0, 1\}$
- $P = \{p_1, p_2, ..., p_r\}$ is the set of actions' probability
- *T* is the learning algorithm
- $p(n+1) = T[p(n), α(n), β(n)]$ is the reinforcement scheme which could be linear if $p(n+1)$ is a linear function of $p(n)$, otherwise it is nonlinear.

In $n_{th}$ step of VLSA, if the $i_{th}$ selected action $α_i(n)$ receives the reward reinforcement signal $β_i(n) = 0$, its corresponding probability $p_i(n+1)$ is updated using ( 3 ). If it receives the penalty reinforcement signal $β_i(n) = 1$, its corresponding probability $p_i(n+1)$ is updated using ( 4 ).

$$p_j(n+1) = \begin{cases} p_j(n) + a \times (1 - p_j(n)) & j = i \\ p_j(n) \times (1 - a) & \forall j \mid j \neq i \end{cases} \quad (3)$$

$$p_j(n+1) = \begin{cases} p_j(n) \times (1 - b) & j = i \\ b \times (r - 1) + (1 - b) \times p_j(n) & \forall j \mid j \neq i \end{cases} \quad (4)$$

Where in ( 3 ) and ( 4 ), *a* and *b* are learning parameters. Different values for *a* and *b* creates different learning algorithms:

- If a = b, the learning algorithm will be of type of Linear Reward-Penalty ($L_{RP}$)
- If b = 0, the learning algorithm will be of type of Linear Reward-Inaction ($L_{RI}$)
- If a » b, the learning algorithm will be of type of Reward-epsilon-Penalty ($L_{RεP}$)

III. ADAPTIVE INTELLIGENCE OPTIMIZER (AIO)

*A. How to Map PSO in the SIS's Computation Cycles*

The SIS [18] characterizes as a general purpose system consisting of different decision cycles. Each SIS's decision cycle can encompass different learning algorithms. Within each cycle, the SIS lets the learning algorithm to improve its performance via performing a number of SIS operations. Different decision cycles share information through sending candidate solutions to the next cycle. The basic workflow of SIS enables combining different learning algorithms and implicit knowledge sharing. Through a sequence decision cycle, the SIS gradually reaches to a synergy between environment and learning algorithm.

The proposed AIO is a combination of PSO, SIS and LA. We map the current framework of SIS into AIO as follows:

- **Enumeration** (-enum<) consists of calculating the fitness of all available particles. The fitness information will be passed to the next phase for future use.

- **Propagation** (=prop+) in AIO defines as the *personal best position* for each particle (*pbest*) and *global best position* for the entire swarm (*gbest*). This information is calculated in each decision cycle and SIS can utilize this information in order to propagate the current experience to other subcomponents.

- **Adaptation** (+adap=) utilizes the current solutions to produce the elitist next generation of particles. In AIO, the adaptation behavior can be seen when the particles move toward the *pbest* and *gbest*.

- **Elimination** (>elim-) rules out a set of infeasible solutions and keeps the feasible solutions for the next decision cycle. The elimination operator is implemented in the context of AIO by extracting the *k* best generated solutions for the current decisions cycle where *k* is the enumeration factor and lets SIS to extract an elite subset of AIO's solution.

- **Concentration** (>conc=) tries to concentrate on the elite population. In AIO, the concentration step consists of updating the velocity and position of elite particles along with other non-elite particles. But the difference between these two groups of particles will be the fact that the non-elite particles will have a random mutation to see whether their fitness can be improved or not.

SIS consists of a set of super-components. In each computation cycle of AIO, the SIS's operators are used for PSO super-component to produce the results. A multilevel set of computation cycles of AIO is as follows:

$cycle_1$: [$guard_{1,2}$] $P_0$ –enum< $P_1$ =prop+ $P_2$ >elim- $P_3$ >conc= $P_4$ (5)

$cycle_1$: [$guard_{2,1}$] $P_0$ –enum< $P_1$ =prop+ $P_2$ +adap= $P_3$ >elim- $P_4$ >conc= $P_5$ (6)

where in (5), the first computation cycle $cycle_1$ does not produce a new feasible solution which means the *pbest* or *gbest* positions of PSO does not change, whereas in (6) for the second computation cycle $cycle_2$ *pbest* or *gbest* positions are improved and concentration operator is applied to the solutions.

### B. Simulating SIS's slow/quick decision cycles

In AIO, we simulates the slow and quick decision cycles of SIS by changing the step size of the *inertia weight w* parameter of PSO. The slow and quick decision cycles are as follows:

$w = w_{max} - (0.75 * i (w_{max} - w_{min})/i_{max})$ (7)

$w = w_{max} - (i (w_{max} - w_{min})/i_{max})$ (8)

where in (7) and (8), $w$, $w_{max}$, $w_{min}$ are inertia weight, maximum and minimum allowed weights. Also, $i$ and $i_{max}$ are the current iteration and maximum number of iterations.

### C. Incorporating TDR system of SIS in AIO

The Chi (Qi) concept of SIS to break the problem dimensions into three sub-dimensions of Tian, Den, Ren (TDR). These sub-dimensions show different subcomponents of the SIS where:

- **Tian** is the heaven
- **Di** is earth
- **Ren** is the human being interacting with environment.

According to Chinese philosophy, these components are the essential ingredients of a human-centric psycho-physical system. Also, a higher level component called Chi (Qi) can be added to the TDR system to control these subcomponents.

In AIO, we do not just see the problem space as three different sequential sub-dimensions. We organize the problem overly as a set of independent dimensions where subsets of them can be optimized with each other within a super component. In case of the TDR system, the number of super components are 3, but AIO can accept any arbitrary number of super-component *k* starting from 1 to the number of dimensions of the problem.

The concept of super-component in SIS can be translated into the *swarm* keyword in the SI terminology. Considering each subcomponent of SIS as a swarm of particles in the PSO. The PSO population is divided into number of swarms corresponding to the number of SIS's super-component. Each swarm contains a sub-part of the problem space which means a set of arbitrary dimensions can constitute a swarm of particle. Introducing the concept of swarm, how the swarm membership is implemented in AIO.

Each dimension of the PSO population is controlled by a learning automaton which selects the dimension's swarm membership. Figure 1 shows the LA distribution for swarm membership selection over the problem space.

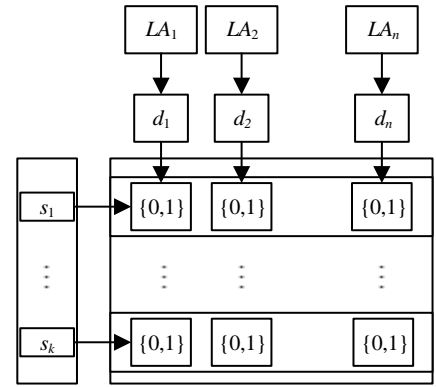

Figure 1 Placing LA on top of problem dimensions to select swarm membership.

where in Figure 1:

- $D = \{d_1, \ldots d_n\}$ is the problem dimensions
- $LA = \{LA_1, \ldots, LA_n\}$ is the set of LA allocated to each dimension
- $S = \{s_1, \ldots, s_k\}$ is the set of swarms

In Figure 1, each dimension is armed with a leaning automaton that decides which swarm is going to be participated in for the next iteration. LA have an action set where 1 means the dimension is going to be a member of a specific swarm.

The next step in AIO is to update the probability vector of LA based on the performance of the swarm. Since each swarm is composed of a subset of the problem dimensions, we use a *context vector* [19] to evaluate swarm's particles where we used the particle value itself for the dimensions that belong to the swarm and use the *gbest* position for the rest of dimensions. In this way, we managed to create an *n* dimensional vector while trying to evaluate the swarm's particles.

While running the AIO, if we face a better solution, we will update the *pbest* and *gbest* positions. The AIO utilizes this information to create the reinforcement signal for LA selecting the swarm membership. Figure 2 shows how LA acts as the SIS' controller while controlling the swarm's super-components.

In Figure 2, $\alpha$ and $\beta$ are LA's action set and reinforcement signal set produced from the PSO population. The LA act as the SIS controller where in each iteration, it try to select the optimal swarm membership and refines this configuration applying the reinforcement signal produced from the environment. Each LA has an action set equal to the number of TDR sub dimensions. In the LA actions selection step, LA select one of the swarm for the corresponding dimension mounted on. Also, in the LA probability update step, LA use the reinforcement signal produced from the PSO to update the probability vectors.

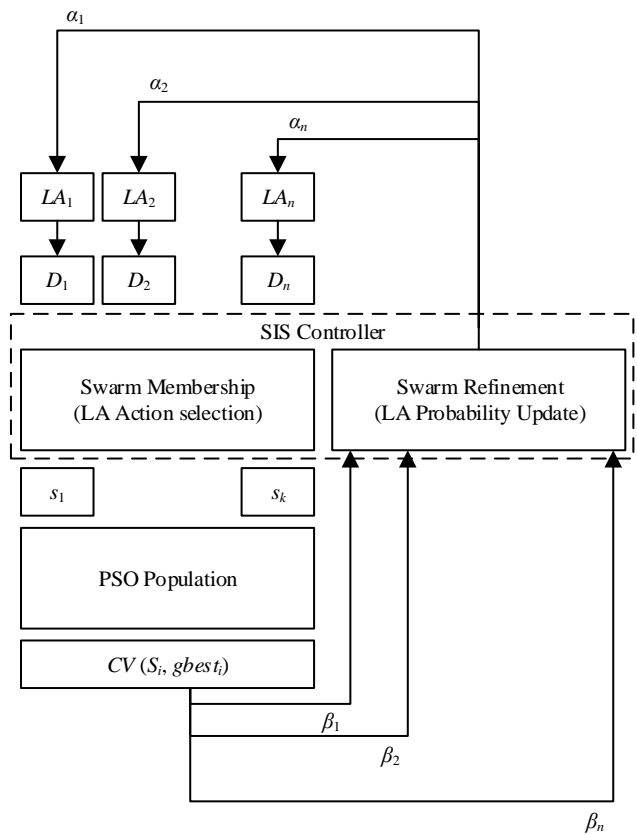

Figure 2 SIS controller and its position within the AIO framework where $i \in \{1, \ldots, k\}$ is the swam index.

### D. Utilizing the Bi-population PSO

AIO uses two isolate PSO populations in order to easily escape from the local minima while getting trap. Figure 3 shows how mounted LA on the swarms select the population for them. In each iteration, these LA select one of the PSO's population for the swarm, then the performance of the swarm is calculated using the selected population via the context vector. Moreover, Figure 3 shows that, in each episode, AIO updates the LA probability vector based on the reinforcement signal that is calculate from the swarm's fitness improvement with regard to *gbest* information. This scheme is similar to what we showed for SIS controller in Figure 2.

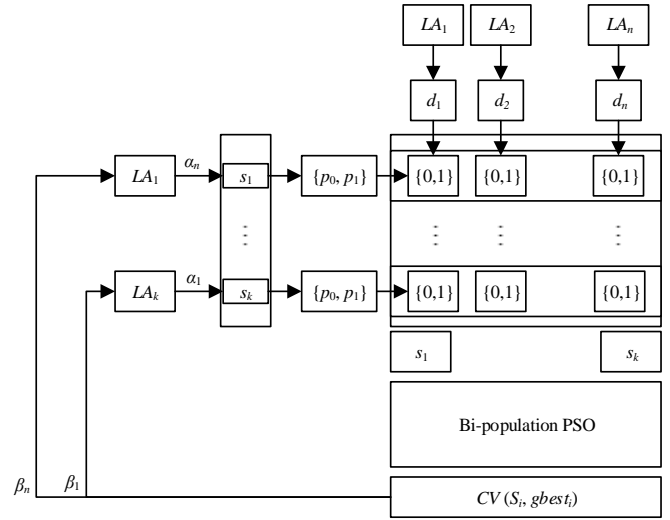

Figure 3 Bi-population PSO incorporated in the context of AIO for optimization isolation.

Figure 3 shows how two sets of LA are incorporated in AIO. The *vertical* set of LA determines the swarm membership and the *horizontal* set of LA selects the population for each swarm. In this way, the AIO learns in two dimensions where, it vertically learn how to pick the best configuration for the swarms and horizontally learns which population fits for a specific swarm.

## IV. EXPERIMENTS

### A. Benchmark Fucntions

In order to evaluate the performance of AIO, we pick 5 different benchmarks including:

- **Sphere** ( 9 ) is a continues, convex, and unimodal benchmark function which has a bowl shaped structure and a single global minimum.

$$f(x) = \sum_{i=1}^{n} x_i^2 \qquad (9)$$

- **Rosenbrock** ( 10 ) is a unimodal function with a unique global minimum lies in a narrow, parabolic valley. Even though, finding the valley is not very hard, convergence to the minimum is difficult.

$$f(x) = \sum_{i=1}^{n-1} \left(100(x_{i+1} - x_i^2)^2 + (x_i - 1)^2\right) \quad (10)$$

- **Ackley** ( 11 ) is a continues, differentiable, non-separable, and multi-modal benchmark function which is characterized by a nearly flat outer surface and a large void hole at the center.

$$f(x) = -a \exp\left(-0.2\sqrt{\frac{1}{n}\sum_{i=1}^{n} x_i^2}\right) \\ -\exp\left(\frac{1}{n}\sum_{i=1}^{n}\cos(2\pi x_i)\right) + 20 + \exp(1) \quad (11)$$

- **Griewanks** ( 12 ) is a continues, differentiable, non-separable, and multi-modal benchmark which has many regularly distributed widespread local minima.

$$f(x) = \sum_{i=1}^{n} \frac{x_i^2}{4000} - \prod_{i=1}^{n} \cos\left(\frac{x_i}{\sqrt{i}}\right) + 1 \quad (12)$$

- **Rastrigin** ( 13 ) is a continues, differentiable, and multi-modal benchmark function with several regularly distributed local minima.

$$f(x) = 10n + \sum_{i=1}^{n} \left(x_i^2 - 10\cos(2\pi x_i)\right) \quad (13)$$

### B. Experimental Setup

The AIO and its counterpart algorithm PSO are implemented in Python 3.4 and the code is available at https://github.com/hmofrad/pso. The followings are the configurations of AIO and PSO:

- Maximum number of iterations $i_{max}$ is 10,000
- $c_1$, and $c_2$ are set to 1.49445
- $w$ is set to 0.74 for PSO
- $w_{max}$ and $w_{min}$ are set to 0.9 and 0.4 for AIO
- Number of dimensions is 30 for both algorithms
- Population size is 50 for both algorithms
- Elite factor for AIO is 2/3
- TDR factor for AIO is 5
- α = β = 0.1 for LA

### C. Implementing an XML Interface

An XML interface is implemented for both AIO and PSO algorithm that contains a specification file. In the root tag of this XML interface, we define the optimization algorithm which is PSO as a type of the super-components of AIO algorithm. Moreover, in the leaf tags of XML interface, we define different parameters of the algorithm. These parameters are introduced in the previous section. The AIO implementation easily lets new super-component to add in the specification file. One can change PSO with any SI algorithm e.g. Genetic Algorithm (GA) and adds its specification to the XML interface.

### D. Experiments

The AIO and PSO results are shown in Figure 4 – Figure 8. The horizontal axis of these figures denotes the iteration number and vertical axis represents recorded average fitness evaluation value corresponding to the iteration number over 5 runs.

Figure 4 is the results for the Sphere function. Sphere is an easy, unimodal function. AIO's performance surpasses PSO because of utilizing SIS framework and basically benefiting from the controller component.

Figure 5 shows the results for Rosenbrock function. AIO obtains a remarkable performance while optimizing this function. The two PSO populations of AIO help the algorithm scape from the local minima which is shown in the first half of the figure where the AIO stagnates a little until iteration 5,000.

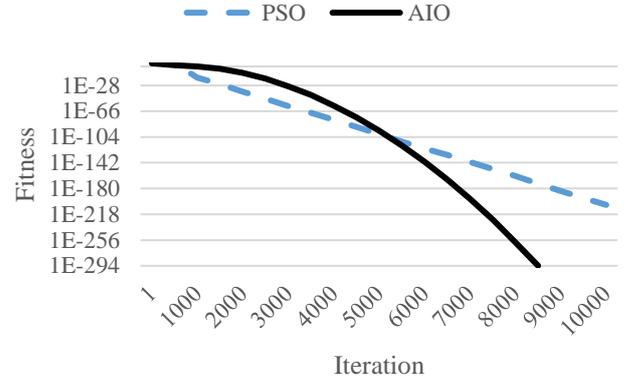

Figure 4 Results of PSO and AIO for a 30 dimensional *Sphere* benchmark.

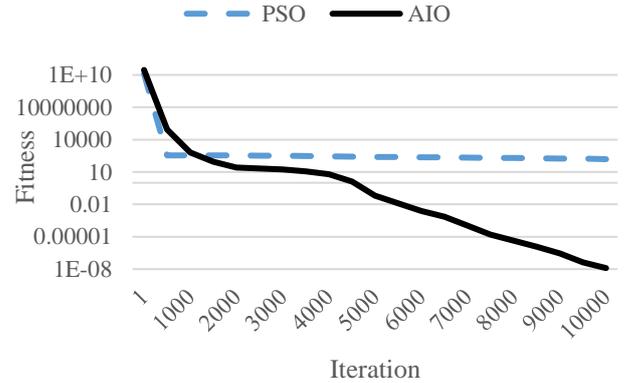

Figure 5 Results of PSO and AIO for a 30 dimensional *Rosenbrock* benchmark.

From Figure 6, AIO outperforms PSO with a huge difference. It gains this performance because of a collective selection of swarm membership and dividing the problem space into a set of sub-problem optimizing them individually.

Figure 7 shows the performance of 2 algorithms for Griewanks function. Since Griewank is a complex multimodal function, the AIO cannot save its high performance margin compare to previous benchmarks but still it gains positive a little bit of performance advantage compare to PSO.

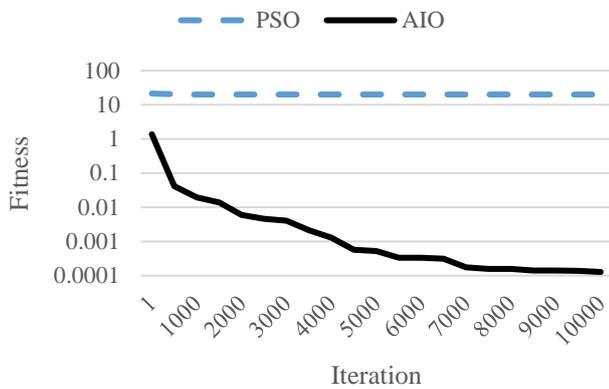

Figure 6 Results of PSO and AIO for 30 dimensional *Ackley* benchmark.

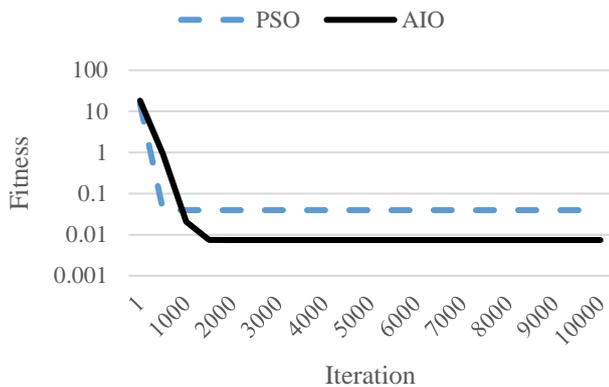

Figure 7 Results of PSO and AIO for a 30 dimensional *Griewanks* benchmark.

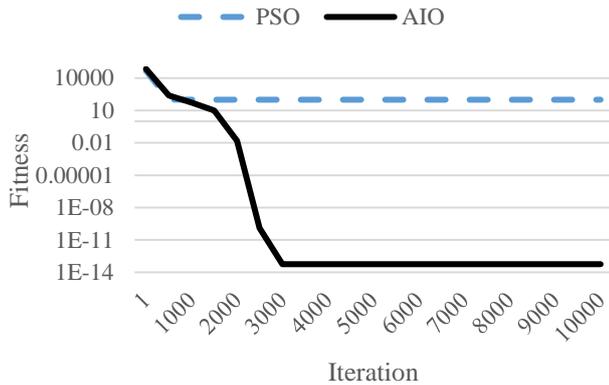

Figure 8 Results of PSO and AIO for a 30 dimensional *Rastrigin* benchmark.

Figure 8 shows the results for Rastrigin benchmark. From this figure, one can see AIO tries to adaptively incorporate its two population in order to escape from the local minima during the first 3,000 iterations but finally it halts after this iteration.

Finally, we have tested the PSO and AIO on 5 different optimization benchmark functions. The AIO outperforms PSO in all benchmarks showing that an EA combining with the compartmentalize framework of SIS and adaptive framework of LA gains a considerable performance improvement.

## V. FUTURE WORK

The AIO is a fast yet simple optimization algorithm designed for balancing the exploration and exploitation power of SI. The AIO can have applications in parameter tuning of ML models like Artificial Neural Network (ANN) where it iteratively finds the optimal input parameters of an ANN through a sequence of modeling the data, calculating the performance of the model, and giving the feedback to AIO on how good the selected parameter configuration is. Moreover, one can use AIO for feature selection, dimensionality reduction, or clustering where we want to shrink the problem space or put correlated dimensions within a same bucket.

## VI. CONCLUSION

In this paper, we introduce AIO which is an adaptive bi-population optimizer implemented in the context of SIS. The subcomponents of AIO are swarms of particle which are assembled using LA. Also, the population for each of these swarms is selected using another set of LA. The AIO collectively tries to optimize a problem in two dimensions: It optimizes the swarm membership configuration vertically which is basically a type of dimensionality clustering task and it horizontally finds a preferred population to optimize the selected dimensions. The AIO is tested on 5 unimodal and multimodal benchmark functions and the results show that it significantly overcomes the standard PSO in all benchmark functions.

REFERENCES

[1] J. Kennedy, "Swarm intelligence," *Handb. Nat.-Inspired Innov. Comput.*, pp. 187–219, 2006.
[2] J. Kennedy and R. Eberhart, "Particle swarm optimization," in , *IEEE International Conference on Neural Networks, 1995. Proceedings*, 1995, vol. 4, pp. 1942–1948.
[3] S.-K. Chang, "Slow intelligence systems," in *International Conference on Multimedia Modeling*, 2010, pp. 1–1.
[4] K. S. Narendra and M. Thathachar, "Learning Automata : A Survey," *Syst. Man Cybern. IEEE Trans. On*, no. 4, pp. 323–334, 1974.
[5] M. Hasanzadeh, M. R. Meybodi, and S. S. Ghidary, "Improving Learning Automata based Particle Swarm: An optimization algorithm," in *2011 IEEE 12th International Symposium on Computational Intelligence and Informatics (CINTI)*, 2011, pp. 291–296.
[6] M. Hasanzadeh, M. R. Meybodi, and M. M. Ebadzadeh, "A robust heuristic algorithm for Cooperative Particle Swarm Optimizer: A Learning Automata approach," in *2012 20th Iranian Conference on Electrical Engineering (ICEE)*, 2012, pp. 656–661.
[7] M. Hasanzadeh, M. R. Meybodi, and M. M. Ebadzadeh, "Adaptive Parameter Selection in Comprehensive Learning Particle Swarm Optimizer," in *Artificial Intelligence and Signal Processing*, 2013, pp. 267–276.
[8] M. Hasanzadeh-Mofrad and A. Rezvanian, "Learning Automata Clustering," *J. Comput. Sci.*, vol. 24, pp. 379–388, Jan. 2018.
[9] M. Hasanzadeh, M. R. Meybodi, and M. M. Ebadzadeh, "Adaptive cooperative particle swarm optimizer," *Appl. Intell.*, vol. 39, no. 2, pp. 397–420, Sep. 2013.
[10] M. Hasanzadeh, S. Sadeghi, A. Rezvanian, and M. R. Meybodi, "Success rate group search optimiser," *J. Exp. Theor. Artif. Intell.*, vol. 0, no. 0, pp. 1–17, Nov. 2014.
[11] M. Hasanzadeh and M. R. Meybodi, "Grid resource discovery based on distributed learning automata," *Computing*, vol. 96, no. 9, pp. 909–922, Sep. 2014.
[12] M. Hasanzadeh and M. R. Meybodi, "Distributed optimization Grid resource discovery," *J. Supercomput.*, vol. 71, no. 1, pp. 87–120, Jan. 2015.
[13] M. Hasanzadeh Mofrad, S. Sadeghi, A. Rezvanian, and M. R. Meybodi, "Cellular edge detection: Combining cellular automata and cellular

learning automata," *AEU - Int. J. Electron. Commun.*, vol. 69, no. 9, pp. 1282–1290, Sep. 2015.

[14] Y. Tang, H. Zhang, Z. Liang, and S.-K. Chang, "Social Network Models for the TDR System."

[15] J. Kennedy, "Particle swarm optimization," in *Encyclopedia of machine learning*, Springer, 2011, pp. 760–766.

[16] S.-K. Chang, "A general framework for slow intelligence systems," *Int. J. Softw. Eng. Knowl. Eng.*, vol. 20, no. 1, pp. 1–15, 2010.

[17] M. Thathachar and P. S. Sastry, "Varieties of learning automata: an overview," *Syst. Man Cybern. Part B Cybern. IEEE Trans. On*, vol. 32, no. 6, pp. 711–722, 2002.

[18] S.-K. Chang, Y. Wang, and Y. Sun, "Visual Specification of Component-based Slow Intelligence Systems.," in *SEKE*, 2011, pp. 1–8.

[19] F. van den Bergh and A. P. Engelbrecht, "A Cooperative approach to particle swarm optimization," *IEEE Trans. Evol. Comput.*, vol. 8, no. 3, pp. 225–239, Jun. 2004.